%% file: main.tex
\documentclass[journal=jacsat,manuscript=article]{achemso}

\usepackage[version=3]{mhchem} 
\usepackage[noabbrev]{cleveref}
\usepackage{xspace}
\usepackage{algorithm}
\usepackage{algpseudocode}
\usepackage{graphicx}
\usepackage{caption}
\usepackage{subcaption}
\usepackage{multirow}
\usepackage{makecell}
\usepackage{booktabs}
\usepackage{xcolor}
\usepackage{enumerate}
\usepackage{xurl}
\usepackage{threeparttable}

\input{math_commands}

\input{constants}

\author{Yujie Qian}
\email{yujieq@csail.mit.edu}
\author{Jiang Guo}
\affiliation[csail]
{Computer Science and Artificial Intelligence Laboratory, MIT, Cambridge, MA, 02139}
\author{Zhengkai Tu}
\affiliation[cheme]
{Department of Chemical Engineering, MIT, Cambridge, MA, 02139}
\author{Zhening Li}
\affiliation[csail]
{Computer Science and Artificial Intelligence Laboratory, MIT, Cambridge, MA, 02139}
\author{Connor W. Coley}
\affiliation[cheme]
{Department of Chemical Engineering, MIT, Cambridge, MA, 02139}
\author{Regina Barzilay}
\email{regina@csail.mit.edu}
\affiliation[csail]
{Computer Science and Artificial Intelligence Laboratory, MIT, Cambridge, MA, 02139}

\title{\ours: Robust Molecular Structure Recognition with Image-To-Graph Generation}

\abbreviations{IR,NMR,UV}

\begin{document}

\begin{abstract}
Molecular structure recognition is the task of translating a molecular image into its graph structure. Significant variation in drawing styles and conventions exhibited in chemical literature poses a significant challenge for automating this task. In this paper, we propose \ours, a novel image-to-graph generation model that explicitly predicts atoms and bonds, along with their geometric layouts, to construct the molecular structure. Our model flexibly incorporates symbolic chemistry constraints to recognize chirality and expand abbreviated structures.  We further develop data augmentation strategies to enhance the model robustness against domain shifts. In experiments on both synthetic and realistic molecular images, \ours significantly outperforms previous models, achieving 76--93\% accuracy on public benchmarks. Chemists can also easily verify \ours's prediction, informed by its confidence estimation and atom-level alignment with the input image. \ours is publicly available through Python and web interfaces: \url{https://github.com/thomas0809/MolScribe}.
\end{abstract}

\input{intro}
\input{method}
\input{experiment}

\input{conclusion}

\begin{suppinfo}
\rev{
Data processing details, example molecular images in the benchmarks, implementation details of data augmentation, the algorithm for expanding abbreviations, additional evaluation results, and example predictions on hand-drawn molecules are available in the supporting information.
}
\end{suppinfo}

\begin{acknowledgement}

The author thanks the members of Regina Barzilay's Group at MIT CSAIL for helpful discussions and feedback. This work was supported by the DARPA Accelerated Molecular Discovery (AMD) program under contract HR00111920025 and the Machine Learning for Pharmaceutical Discovery and Synthesis Consortium (MLPDS).

\end{acknowledgement}

\bibliography{reference}


\end{document}

%% file: math_commands.tex
\usepackage{amsmath,amsfonts,bm}

\def\eqref#1{equation~\ref{#1}}

\def\1{\bm{1}}

\def\vh{{\bm{h}}}

\DeclareMathAlphabet{\mathsfit}{\encodingdefault}{\sfdefault}{m}{sl}
\SetMathAlphabet{\mathsfit}{bold}{\encodingdefault}{\sfdefault}{bx}{n}



%% file: constants.tex
\newcommand{\ours}{MolScribe\xspace}

\newcommand{\rev}[1]{{#1}}

%% file: intro.tex
\section{Introduction}

\begin{figure}[t]
    \centering
    \includegraphics[width=0.9\linewidth]{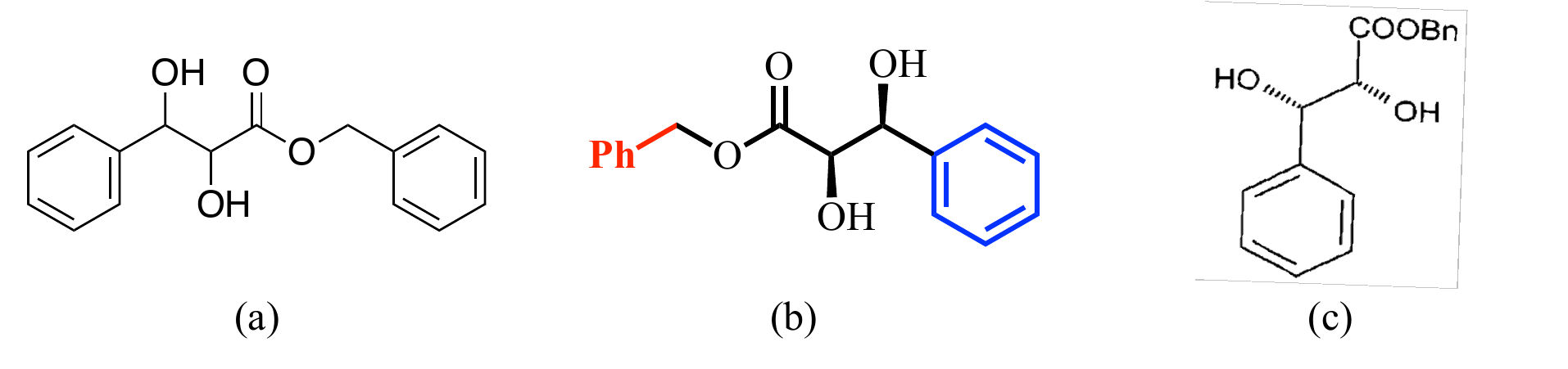}
    \caption{Three images of the same molecule with different layouts and styles. (a) depicts the full molecular structure and does not specify chirality. (b) and (c) use different abbreviations, and (b) color-codes the phenyl groups. }
    \label{fig:intro}
\end{figure}

Molecules are commonly drawn as 2D images in chemistry literature. Such drawings exhibit a wide variety of styles, which complicates the task of translating these images into machine-readable molecular structures.
For example, \Cref{fig:intro} shows three different ways to draw the same molecule. The plain image (a) can be recognized adequately by existing models, but variants such as (b) and (c) are more difficult as they involve \rev{stereochemistry}, functional group abbreviations, and diverse drawing styles. While prior work has demonstrated that it is possible to train models that perform well on one style assuming labeled data is available, the diversity of possible styles is a standing challenge. We cannot assume access to training data that covers all possible styles and patterns in the chemistry literature due to the high annotation cost. Therefore, we aim to enhance the robustness of molecular structure recognition model to generalize to an arbitrary molecular image.

In this paper, we propose \ours, which takes as input a molecular image (e.g. a PNG or JPG file), and generates its molecular graph structure. The model relies on three complementary approaches to handle data variation robustly. First, \ours explicitly predicts the atoms and bonds along with their geometric layouts in the image, which together constitute a \textit{2D molecular graph}. Second, we incorporate chemistry knowledge as symbolic constraints to the model, such that it can accurately recognize complex chemical patterns. For example, we determine the chirality of asymmetric atoms based on the predicted graph and layout, and design an algorithm to parse abbreviated functional groups that commonly appear in molecular images. Third, we propose data augmentation strategies to synthetically generate images that cover diverse drawing styles during training.  Specifically, \ours is implemented as an image-to-graph generation model, which abstracts the input image into hidden representations and generates the molecular graph with an autoregressive decoder. \ours's output can be aligned with the input image at the atom-level, allowing humans to easily interpret its predictions.

\ours combines the strengths of previous rule-based and neural models. Rule-based methods\cite{osra,molvec} can easily enforce chemistry constraints and conduct symbolic reasoning over the molecular graph, while neural models~\cite{mse-staker,img2mol,image2graph} are more robust to image styles and perturbations. \ours naturally extends a neural generation model to predict atoms and their coordinates as a sequence, and then the bond between each pair of atoms. This design allows \ours to robustly recognize local atoms and bonds and flexibly integrate chemistry rules when constructing the molecular graph, such that it generalizes to different scenarios.

\rev{In our experiments, \ours outperforms an image-to-SMILES baseline on both in-domain and out-of-domain images, and achieves strong recognition accuracy (76--93\%) on five public benchmarks.} We also construct a new benchmark with molecular images from journal publications, and \ours significantly outperforms the baseline and existing systems. Moreover, \ours is robust against input perturbation and low-quality images, predicts chirality more accurately, and provides confidence estimation. Finally, we conduct a human evaluation to understand how our model can help chemists to parse molecular images in a semi-automated workflow. The results show that our graph prediction significantly reduces the time needed by chemists to extract the molecular structures from images. Our model, code, and data are publicly available (\url{https://github.com/thomas0809/MolScribe}) for future research in molecular structure recognition and chemistry information extraction.  We have developed a demo that allows chemists to use \ours from a web interface (\url{https://huggingface.co/spaces/yujieq/MolScribe}).

%% file: method.tex
\section{Molecular Structure Recognition}


\subsection{Background}
Research on molecular structure recognition dates back to at least the 1990s (see a recent review by \citeauthor{rajan2020review}~\cite{rajan2020review}). \rev{The task is also known as Optical Chemical Structure Recognition (OCSR).} Earlier rule-based systems~\cite{mcdaniel1992kekule,casey1993optical,ibison1993chemical,valko2009clide,park2009automated,sadawi2012chemical,algorri2007reconstruction,frasconi2014markov} relied on traditional image processing techniques (binarization, line smoothing and thinning, vectorization) to segment the pixels into atoms and bonds, and standalone optical character recognition (OCR) models to identify atom labels. Then, heuristics based on line length, width, spacing, and direction were used to determine bond types (single, double, triple, wedge, dashed) and connect all these elements into molecular graphs. These systems were usually meticulously engineered by chemists, with specific rules to handle different situations in molecular images. For example, they usually have an expert-compiled dictionary to resolve the most common functional group abbreviations. A few open-source tools have been developed~\cite{osra,smolov2011imago,molvec}, and developers are continuously improving them by writing new rules to cover edge cases (e.g., bridge bonds)\cite{osra}. The rule-based systems have achieved decent recognition accuracy on patent images, but there is still much room for improvement on journal article images, which are more diverse. Researchers also identified that small image perturbations can cause significant performance degradation~\cite{img2mol}.

\begin{figure}[t]
    \centering
    \includegraphics[width=0.45\linewidth]{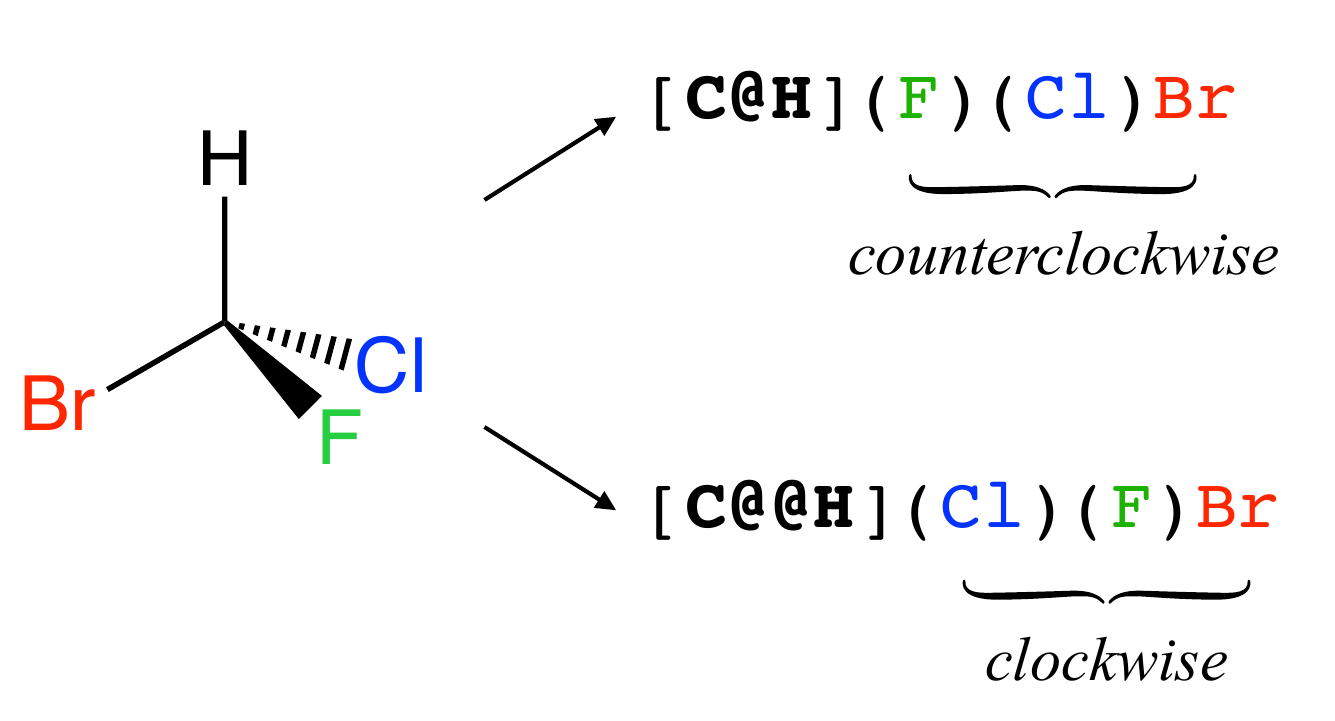}
    \caption{Chirality specification in SMILES. A chiral center may be indicated by ``@'' or ``@@'', meaning the neighbors are listed counterclockwise or clockwise, in this example when looking down the H-C bond.}
    \label{fig:chiral}
\end{figure}

Neural models have been proposed to improve the robustness against image variations. \citeauthor{mse-staker} presented an image-to-SMILES generation model~\cite{mse-staker}: a convolutional neural network encodes the input molecular image into a hidden representation, and a recurrent neural network decodes the SMILES string of the molecule. \rev{A few variants have been proposed since, exploring multiple model architectures, such as Inception network\cite{rajan2020decimer}, Transformer\cite{rajan2021decimer,image2smiles}, Swin Transformer\cite{swinocsr}, pre-trained decoder\cite{img2mol}, graph decoder\cite{image2graph}, and also the application to hand-drawn molecules~\cite{D1SC02957F,decimer_hdm}.} Neural models enjoy the simplicity of end-to-end training and the strength of handling different image styles, but operating on the SMILE strings instead of explicitly recognizing atoms and bonds makes it difficult to incorporate chemistry constraints.
For example, as we show in the experiments, \rev{predicting stereochemistry presents a challenge for traditional neural networks due to the need for geometric reasoning over the molecular graph.} \Cref{fig:chiral} shows two equivalent SMILES strings for a chiral molecule. Whether the chiral center should be written as ``[C@H]'' or ``[C@@H]'' depends on the relative order of its connecting bonds, which cannot be easily determined from the local pattern. Due to the lack of explicit notion of atoms and bonds, \rev{we cannot easily inject chemistry rules into these models.} 



Besides the image-to-SMILES formulation that directly generates the SMILES string, ChemGrapher~\cite{chemgrapher} and MolMiner~\cite{molminer} train separate modules to detect atoms, bonds and texts, based on image segmentation or object detection; and then construct the graph with heuristics. Such systems can also incorporate chemistry constraints during their graph construction process. Compared to these models, we simplify the method with a single end-to-end model to generate the molecular graph, so that our model does not rely on heuristics to connect the local predictions.


\subsection{Task Formulation}

Molecular structure recognition is the task of translating single-molecule images into corresponding molecular structures. 
In this paper, we formulate it as \textit{image-to-graph generation}. Given an image $I$ of molecule $M$, we translate it into a {2D molecular graph} $G=(A, B)$, where $A=\{a_1, a_2, \dots, a_n\}$ is the set of atoms, $B\subset A\times A\times T$ is the set of bonds, and $T$ is the set of bond types (e.g.\ single, double, triple, solid wedge, dashed wedge). We define each atom as $a_i=(l_i, x_i, y_i)$, where $l_i$ is the atom's corresponding SMILES (sub)string (e.g. ``$\mathrm{C}$'', ``$\mathrm{N}$'', ``$\mathrm{[OH-]}$'', and ``$\mathrm{[235U]}$''), and $x_i$ and $y_i$ are the 2D coordinates of the atom in the image.
Finally, the molecular graph can be converted and stored in standard data formats, such as a SMILES string~\cite{smiles} or a MOLfile~\cite{chem_file_format}.


\subsection{Model}
\begin{figure}[t]
    \centering
    \includegraphics[width=\linewidth]{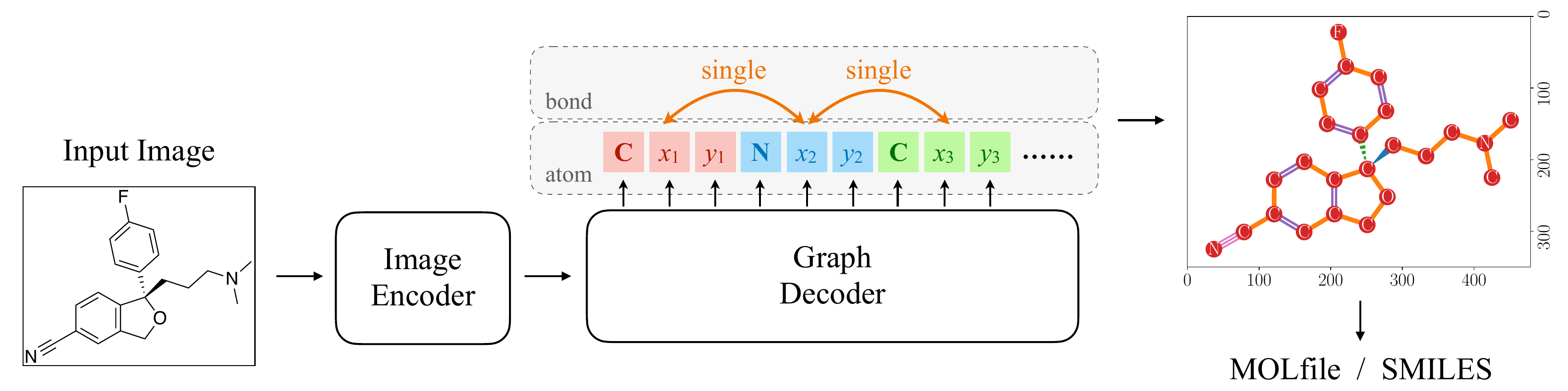}
    \caption{\ours model architecture. The input image is encoded with an image encoder, and the graph decoder predicts the atoms and bonds. A molecular graph is constructed from the predictions, and is converted to a MOLfile or a SMILES string.}
    \label{fig:molmodel}
\end{figure}

We propose \ours, an image-to-graph model that translates an input image into a 2D molecular graph. The model follows an encoder-decoder architecture, as shown in \Cref{fig:molmodel}. We first use an image encoder to encode the input image into a hidden representation, and then use a graph decoder to generate the molecular structure. Unlike traditional image-to-SMILES models, which output a sequence of tokens, our decoder predicts atoms, bonds, and their geometric layouts jointly, such that the molecular structure can be reconstructed. 

\ours formulates image-to-graph translation as a conditional generation process:
\begin{equation}
    P(G\mid I) = P(A\mid I)~P(B\mid A, I) 
    \label{eq:graph}
\end{equation}
where $P(A\mid I)$ and $P(B\mid A, I)$ are parametrized as an atom predictor and a bond predictor, respectively.

As shown in the middle of \Cref{fig:molmodel}, the atom predictor is an autoregressive decoder that generates the atoms in a sequence, i.e.,
\begin{equation}
    P(A\mid I) = \prod_{i=1}^n P(a_i \mid A_{<i}, I)
\end{equation}
where $A_{<i}$ stands for the atoms before $a_i$. Inspired by the object detection model Pix2Seq~\cite{pix2seq}, our atom predictor simultaneously predicts atom labels and coordinates. Specifically, we construct a sequence of discrete tokens as the output format of the atom predictor, 
\begin{equation}
    S^A = [l_1, \hat{x}_1, \hat{y}_1, l_2, \hat{x}_2, \hat{y}_2, \cdots, l_n, \hat{x}_n, \hat{y}_n]
\end{equation}
where each atom $a_i$ corresponds to three tokens $l_i$, $\hat{x}_i$ and $\hat{y}_i$. $l_i$ is the atom's SMILES specification, including the element identity, isotope, formal charge, and implicit hydrogen count. $\hat{x}_i$ and $\hat{y}_i$ are discrete values representing the atom's coordinates defined by binning, i.e.,
\begin{equation}
    \hat{x}_i= \left\lfloor \frac{x_i}{W} \times n_\text{bins} \right\rfloor, \quad \hat{y}_i = \left\lfloor \frac{y_i}{H} \times n_\text{bins} \right\rfloor
\end{equation}
where $H$ and $W$ are the height and width of the input image, respectively, and $n_\text{bins}$ is a hyperparameter for the number of bins. This discretization is designed to simplify the model architecture. As the sequence $S^A$ contains all the atom labels and coordinates, we implement the atom predictor to generate it autoregressively,
\begin{equation}
    P(A\mid I) = P(S^A\mid I) = \prod_{i=1}^{|S^A|} P(S^A_i\mid S^A_{<i}, I).
\end{equation}
In practice, we split the molecule's SMILES string with an atom-wise tokenizer~\cite{atomwise_tokenizer} into atom labels and append the coordinate tokens to them. The other tokens in the SMILES string, such as parentheses, bond symbols and digits, indicate connections among the atoms and are helpful to the model, so we keep them in the output sequence $S^A$. These tokens are not associated with coordinates. 

The bond predictor is a feedforward network that predicts the bond between each pair of atoms (see \Cref{fig:molmodel}). Each atom $a_i$ is represented as a vector $\vh_{a_i}$, the hidden state of its last token in the decoder output. For each pair $a_i$ and $a_j$, we concatenate their representations to form the input to the bond predictor, which classifies the bond type. \rev{Formally, 
\begin{equation}
    P(B\mid A, I) = \prod_{i=1}^n \prod_{j=1}^n P(b_{i,j}\mid A, I)
\end{equation}
where $b_{i,j}$ indicates the bond between $a_i$ and $a_j$. We use ``None'' to indicate no bond exists between the atom pair, and other possible bond types include single, double, triple, aromatic, solid wedge and dashed wedge. For single, double, triple, and aromatic bonds, it is expected that $b_{i,j} = b_{j,i}$, but wedge bonds are not symmetric.
To account for this, we predict both directions $b_{i,j}$ and $b_{j,i}$ independently, and merge their predicted probabilities at inference time. For symmetric bonds, $b_{i,j}$ and $b_{j,i}$ are expected to be the same, so the probabilities are averaged, i.e.
\begin{equation}
\hat{P}(b_{i,j}=t) = \frac{1}{2} \big( P(b_{i,j}=t) + P(b_{j,i}=t) \big), \quad 
t \in\{\text{``single'', ``double'', ``triple'', ``aromatic''}\}.
\end{equation}
For asymmetric bonds (wedges), as a solid wedge is equivalent to a dashed wedge in the opposite direction, we set $b_{i,j}=$ ``solid-wedge'' and $b_{j,i}=$ ``dashed-wedge'' if there is a solid wedge from $a_i$ to $a_j$, and vice versa. At inference time, we merge the probabilities by 
\begin{equation}
\begin{split}
&\hat{P}(b_{i,j}=\text{``solid-wedge''}) = \frac{1}{2} \big( {P(b_{i,j}=\text{``solid-wedge''}) + P(b_{j,i}=\text{``dashed-wedge''})} \big), \\
&\hat{P}(b_{i,j}=\text{``dashed-wedge''}) = \frac{1}{2} \big( {P(b_{i,j}=\text{``dashed-wedge''}) + P(b_{j,i}=\text{``solid-wedge''})} \big).
\end{split}
\end{equation}
$\hat{P}(b_{i,j})$ is the bond predictor's final prediction during inference. (The conditional parts are omitted.)
}

Finally, a molecular graph is constructed from the predicted atoms and bonds. We use the RDKit~\cite{rdkit} toolkit to save the molecular structure as a MOLfile (with 2D coordinate information) or a SMILES string (without 2D coordinate information).

\subsubsection{Stereochemistry}
\rev{As explained in \Cref{fig:chiral}, it can be challenging for a neural model to recognize stereochemistry.} In SMILES, chirality is indicated as an atomic property, i.e., $@$ or $@@$ after an atom label specifies the relative spatial orientation of the bonds connected to this atom in the order they are listed. There is no direct correspondence between the local image pattern and whether the atom symbol is $@$ or $@@$. \rev{Understanding stereochemistry requires geometric reasoning, which is not a strength of traditional neural networks\cite{battaglia2018relational}}. In our model, we apply chemistry rules to explicitly determine the stereochemistry, including both chirality for atoms and cis/trans isomerism for double bonds, based on the predicted molecular graph and coordinates. For example, to predict chirality, we first find the bonds connected to each chiral center, infer their relative order by the predicted atom coordinates and determine the chiral type explicitly (with RDKit's implementation). Since the atoms and bonds are easier to predict from the image, this rule-based approach recognizes stereochemistry more accurately than a vanilla neural model.

\subsubsection{Abbreviated Structures}
Another challenge in molecular structure recognition is the parsing of abbreviated structures. Chemists often simplify their drawings of molecules by using abbreviations or condensed formulas, such as ``Me'' for methyl, ``Et'' for ethyl, and ``CHO'' for formyl. The model cannot understand their underlying structures without external knowledge. The abbreviated structures can be treated as ``superatoms'' in the molecular graph, but additional processing is required if we want to derive the complete molecular structure. 
Both rule-based and machine learning models in previous works typically compile a list of common abbreviations and their functional group structures, and substitute the superatoms at inference time. Given the combinatorial nature of abbreviations, this solution may not be sufficient. There is a large amount of combinations of element symbols and functional group abbreviations, e.g., ``OMe'', ``CO$_2$Et'', and ``P(O)(OEt)$_2$''. While they can all be considered as superatoms, it is infeasible to enumerate all possible combinations in a list.

\input{algorithm}

We propose a more flexible method to parse abbreviated structures. We first split the superatom symbols into characters, such that our model is not limited to the prediction space of patterns in the training data and can generalize to unseen patterns.  During inference, we design a greedy algorithm to expand the abbreviated structures. This task is non-trivial because the abbreviations do not follow a rigorous grammar (for example, a carboxyl group can be written as either ``COOH'' or ``CO$_2$H''), and there is no deterministic way to parse them. In \Cref{algo:condensed}, we derive the list of atoms and greedily connect them until their valences are full, based on assumptions of how these abbreviations are typically written to reflect the connections between atoms. 
We adopt this simple solution instead of using another learned model, as our algorithm has already addressed most abbreviations we observe and is extensible.
More details are available in the supporting information.

\subsection{Data}
We train \ours on publicly available data, following previous works~\cite{mse-staker,image2graph}. Our training data comes from two sources:
\begin{itemize}
    \item \textbf{Synthetic data}. \rev{We collect 1 million molecules from the PubChem database~\cite{pubchem}} and automatically render their images using the Indigo toolkit~\cite{indigo}. Atom labels and bond types can be easily obtained from the toolkit, and we modify the source code to access atom coordinates. \rev{ Previous works have combined other toolkits to generate a diverse set of training data\cite{randepict,img2mol}, but many of them do not provide atom coordinates. In this work, the molecules are randomly sampled from PubChem without special constraints. More advanced sampling strategies\cite{qsar.200290002} could be used to ensure a diverse coverage of the chemical space. We encourage future work to analyze this problem. }
    \item \textbf{Patent data}. \rev{We collect 680K examples from patent grants released by the United States Patent and Trademark Office (USPTO)~\cite{uspto_grant}, which contain molecular images and structure labels. As some of our benchmarks were collected from the same source, we do not include the images from the same patents as those in the benchmarks.} This dataset is noisy and exact atom coordinates in the image are not available. \rev{We use the relative coordinates that are available in the MOLfiles and normalize them according to the image size. Details about data processing and examples of the noise in this data can be found in the supporting information. }
\end{itemize}

\subsubsection{Data Augmentation}
We design data augmentation strategies for both molecules and images, so that the training data can cover diverse chemical patterns and drawing styles. 

\begin{figure}[t]
    \centering
    \includegraphics[width=0.6\linewidth]{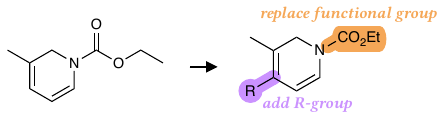}
    \caption{Examples of replacing a functional group with its abbreviation and adding an R-group during synthetic data generation.}
    \label{fig:molaug}
\end{figure}

\paragraph{Functional Groups and R-Groups} 
Synthetic images generated by Indigo never contain functional group abbreviations or R-groups. We dynamically augment the molecules to cover such patterns. \Cref{fig:molaug} illustrates the augmentation process. For functional groups, we construct a list of common abbreviations, where each item consists of a SMARTS pattern for the functional group, an abbreviated label, and a substitution probability. During training, if a functional group exists in a molecule, we randomly replace it with the abbreviation according to the substitution probability. Specifically, the functional group branch is removed from the molecular graph, and a \textit{superatom} with the abbreviation label is added. For R-groups, we also have a list of common R-group labels (R, R$_1$, R$_2$, R', etc.), and randomly add R-groups as superatoms to the molecule. Furthermore, in order to generalize to abbreviations that are not covered in these lists, we add superatoms consisting of random characters. Although they are not necessarily chemically meaningful, this augmentation helps the model acquire the capability of optical character recognition (OCR), so that the model can recognize unseen labels. It does not hurt the model, as the molecular images should be valid at inference time. The functional group and R-group superatoms are associated with labels and coordinates, so our model predicts them in the same way as the other atoms. 

The data augmentation allows our model to handle basic Markush structures~\cite{lynch1996sheffield}, e.g., an R-group attached to a particular atom or within a ring. However, positional variation (a substituent that could be attached to any atom within a ring) and frequency variation (brackets and variables indicating repetition of a substructure) are not covered. Such structures can not be represented in SMILES strings or MOLfiles, thus we leave them as future work.

\paragraph{Drawing styles}
We use image augmentation to improve the robustness of our model with respect to drawing styles. Specifically, we vary the rendering options from Indigo when generating the synthetic data, such as font, bond width, bond length, etc. Furthermore, we apply random perturbations to the molecular images, such as rotation, padding, cropping, and Gaussian noise. Details about the augmentation strategy are available in the supporting information. The image augmentation strategy guarantees our model is trained with diverse image styles and quality, such that it can generalize better in the real world.





%% file: algorithm.tex
{
\hfill
\begin{minipage}{.7\linewidth}
\begin{algorithm}[H]
\caption{Expand Abbreviated Structures} \label{algo:condensed}
\begin{algorithmic}
\algrenewcommand{\algorithmicrequire}{ \textbf{Input:}}
\algrenewcommand{\algorithmicensure}{ \textbf{Output:}}
\Require $G$ --- molecular graph;  
\Require $f$ --- abbreviated structure formula
\Require $a_s$ --- the superatom.

\State $L \gets \mathtt{expand\_subscripts}(f)$ \Comment{e.g.\ $\mathrm{CO_2CH_3 \to COOCHHH}$}
\State $a_\text{cur} \gets L$\texttt{[0]} \Comment{current atom}
\State $G_s.\mathtt{add\_atom}(a_\text{cur})$
\For{$a$ \textbf{in} $L$\texttt{[1:]}}
    \If{$a_\text{cur}.\text{implicit\_valence} = 0$}
        \State \Return FAILURE
    \EndIf
    \State $G.\mathtt{add\_atom}(a)$
    \If{$a.\text{valence} \leq a_\text{cur}.\text{implicit\_valence}$}
        \State $G_s.\mathtt{add\_bond}(a_\text{cur}, a, \text{order}=a.\text{valence})$
    \Else
        \State $G_s.\mathtt{add\_bond}(a_\text{cur}, a, \text{order}=a_\text{cur}.\text{implicit\_valence})$
        \State $a_\text{cur} \gets a$
    \EndIf
\EndFor
\State $G.\mathtt{replace}$($a_s$, $G_s$)
\end{algorithmic}
\end{algorithm}
\end{minipage}
\hfill
\vspace{0.2in}
}



%% file: experiment.tex
\section{Experiments}
\label{sec:exp}

\subsection{Experimental Setup}


\paragraph{Model Implementation}
\label{sec:impl}
Our image encoder is a Swin Transformer~\cite{swin}, a state-of-the-art model in many computer vision tasks. We use the Swin-B model~\cite{swin_github}, which has 88M parameters in total and pre-trained on ImageNet-22K. \rev{A recent work, SwinOCSR\cite{swinocsr}, used the same image encoder as \ours.} We resize the input image to $384\times 384$ resolution for both training and inference. Our decoder is a 6-layer Transformer~\cite{vaswani2017attention} with 8 attention heads, a hidden dimension of 256, and sinusoidal positional encoding. We apply dropout with probability 0.1. 
\rev{The bond predictor is a 2-layer feedforward network with ReLU activation on top of the decoder and has the same hidden dimension.  

The model is trained by teacher forcing, i.e., the decoder is fed with the ground truth token at each step, and predicts the next step conditioned on the previous tokens. The bond predictor takes the hidden states of the decoder as input and predicts the bond between each pair of atoms.  All modules of the model are fully differentiable and trained jointly. (Note that during training, the input to the bond predictor does not depend on the atom predictions.) }
We use a maximum learning rate of 4e-4  with a linear warmup for 5\% steps and a cosine function decay. We use a batch size of 128 and train the model for 30 epochs. We use label smoothing with $\epsilon=0.1$. The atom coordinates are converted to discrete tokens with $n_\text{bin}=64$ in both the $x$ and $y$ directions. During inference, we use greedy decoding to generate the atoms and then predict the bonds. \rev{This procedure is a little different from that during training time, as we have to first autoregressively decode the atoms, and then run the bond predictor.}

\begin{table}[t]
    \centering
    \caption{Summary of the benchmarks for evaluation.}
    \label{tab:data}
    \begin{tabular}{llrr}
    \toprule
    Dataset & Source & \# images & \% chiral \\\midrule
    Indigo & synthetic & 5,719 & 20.2\% \\
    ChemDraw & synthetic & 5,719 & 20.2\% \\\midrule
    CLEF & patent (US) & 992 & 32.7\% \\
    JPO & patent (Japanese) & 450 & 0\% \\
    UOB & catalog & 5,740 & 0\% \\
    USPTO & patent (US) & 5,719 & 20.2\% \\ 
    Staker & patent (US) & 50,000 & 17.3\% \\
    ACS & journal article & 331 & 19.3\% \\
    \bottomrule
    \end{tabular}
\end{table}

\paragraph{Benchmarks}
\Cref{tab:data} lists the benchmarks in our evaluation.
First, we construct two synthetic test sets to evaluate \ours's performance on in-domain and out-of-domain images. The in-domain dataset is generated by Indigo (same as training data), and the out-of-domain dataset is generated by ChemDraw. The two datasets share the same set of 5719 molecules. \rev{We also evaluate our model on five public benchmarks of realistic images: CLEF, JPO, UOB, USPTO~\cite{rajan2020review}, and Staker~\cite{mse-staker}.}

We create a new dataset with 331 molecular images collected from American Chemistry Society (ACS) Publications and manually annotate their SMILES strings. This dataset is a complement to existing benchmarks as they do not contain molecular images from journal articles, which are more diverse in terms of drawing style and use of abbreviations.


\paragraph{Evaluation Metric}
We evaluate the model's recognition performance with exact matching accuracy as the evaluation metric, i.e., the prediction is considered correct only if the entire molecular graph structure matches the ground truth. This metric has been used by previous works \cite{mse-staker,img2mol,image2graph} but with small differences. Specifically, we use RDKit to convert both prediction and ground truth into canonical SMILES, a unique molecular representation, and then compute the string exact match.
Regarding stereochemistry, we require the prediction to match the tetrahedral chirality of the ground truth, but ignore other forms of stereoisomerism (such as cis–trans) because such information is often not available in the ground truth. 
Regarding R-groups, as the their symbols are not allowed in SMILES, we replace them with wildcards (*) during evaluation. For numbered R-group ``R$_d$'' where $d$ is a integer, we replace them with ``$[d*]$''.
We make our evaluation script publicly available, and encourage future work to follow the same metric for fair comparison. \rev{In the supporting information, we present other evaluation metrics, such as accuracy without considering chirality,  and Tanimoto similarity.}


\paragraph{Compared Methods}
We compare \ours with state-of-the-art molecular structure recognition methods. For direct comparison, we train an image-to-SMILES baseline model with the same data and the same encoder-decoder architecture.
We also run existing systems for comparison, including rule-based MolVec~\cite{molvec} and OSRA~\cite{osra}, \rev{and machine learning-based models Img2Mol~\cite{img2mol}, DECIMER (version 2.1.0)~\cite{decimer_v2}, and SwinOCSR~\cite{swinocsr}. } For systems that are not publicly available, such as MSE-DUDL~\cite{mse-staker}, ChemGrapher~\cite{chemgrapher}, and Image2Graph~\cite{image2graph}, we use the reported results in their papers.

\subsection{Results}

\paragraph{Recognition Accuarcy}
\Cref{tab:main} shows the recognition accuracy of different models. Our model \ours achieves consistently higher scores than existing systems on most benchmarks, demonstrating its robust performance in molecular structure recognition. \ours outperforms the baseline image-to-SMILES model, validating the benefits of our graph generation formulation and the integration of symbolic chemistry knowledge. \rev{The JPO dataset involves Japanese characters, which are sometimes recognized by \ours as extra fragments. We apply a post-processing step to keep the largest molecule if there are multiple fragments in the prediction.} The rule-based systems, MolVec and OSRA, achieve decent performance on CLEF, \rev{JPO,} UOB, and USPTO but degrade severely on Staker, in which the images are blurry and low-resolution. Compared with previous neural models such as MSE-DUDL~\cite{mse-staker}, ChemGrapher~\cite{chemgrapher}, Img2Mol\cite{img2mol}, \rev{DECIMER\cite{decimer_v2}}, and Image2Graph~\cite{image2graph}, \ours achieves stronger performance despite being trained on less data. \rev{(We use 1.68M examples in total, while MSE-DUDL uses 68M, ChemGrapher uses 1.5M, DECIMER uses 400 million, Image2Graph uses 7.1M, and Img2Mol uses 11M.) DECIMER performs slightly better than \ours on UOB. The images in this dataset are close to DECIMER's training distribution and relatively simple (no abbreviations, R-groups, or chirality), and thus a neural model trained with more data works well.}
 

\begin{table*}[t]
\rev{
    \centering
    \caption{Molecule structure recognition accuracy on synthetic, realistic, and perturbed benchmarks (scores are exact matching accuracy in \%).}
    \label{tab:main}
    \small
    \setlength{\tabcolsep}{3pt}
    \resizebox{\linewidth}{!}{%
    \begin{threeparttable}
    \begin{tabular}{lccccccccccccccc}
    \toprule
    & \multicolumn{2}{c}{Synthetic} & & \multicolumn{6}{c}{Realistic} & & \multicolumn{4}{c}{Perturbed} \\\cmidrule{2-3} \cmidrule{5-10} \cmidrule{12-15}
                            & Indigo    & ChemDraw  && CLEF  & JPO   & UOB   & USPTO & Staker & ACS     && CLEF$_p$ & UOB$_p$  & USPTO$_p$ & Staker$_p$ \\\midrule
    MolVec                  & 95.4      & 87.9      && 82.8  & 67.8  & 80.6  & 88.4  & 0.8    & 47.4    && 43.7 & 74.5 & 29.7  & 5.0  \\
    OSRA                    & 95.0      & 87.3      && 84.6  & 55.3  & 78.5  & 87.4  & 0.0    & 55.3    && 11.5 & 68.3 & 4.0   & 4.6  \\
    Img2Mol$^*$             & 58.9      & 46.4      && 18.3  & 16.4  & 68.7  & 26.3  & 17.0   & 23.0    && 21.1 & 74.9 & 29.7  & 51.7 \\ 
    DECIMER                 & 69.6      & 86.1      && 62.7  & 55.2  & \textbf{88.2}  & 41.1  & 40.8   & 46.5    && 70.6 & \textbf{87.3} & 46.4  & 47.9 \\
    SwinOCSR$^{**}$         & 74.0      & 79.6      && 30.0  & 13.8  & 44.9  & 27.9  & --     & 27.5    && 32.2 & --   & --    & --  \\
    MSE-DUDL$^\dagger$      & --        & --        && --    & --    & --    & --    & 77.0   & --      && --   & --   & --    & --   \\
    ChemGrapher$^\dagger$   & --        & --        && --    & --    & 70.6  & --    & --     & --      && --   & --   & --    & --   \\
    Image2Graph$^\dagger$   & --        & --        && 51.7  & 50.3  & 82.9  & 55.1  & --     & --      && --   & --   & --    & --   \\\midrule 
    
    Baseline                & 94.1      & 92.2      && 87.4  & 74.8  & \textbf{88.2}  & 91.5   & 86.1   & 59.8   && 88.0 & 87.1 & 91.4 & \textbf{65.9} \\
    \ours                   & \textbf{97.5} & \textbf{93.8} && \textbf{88.9} & \textbf{76.2}  & 87.9 & \textbf{92.6} & \textbf{86.9} & \textbf{71.9} && \textbf{90.4} & 86.7 & \textbf{92.5} & 65.0 \\
    \bottomrule
    \end{tabular}
    \begin{tablenotes}
    \item $^\dagger$ Results from the original papers; -- means not available.
    \item $^*$ Img2Mol does not predict chirality. Additional evaluation results with chirality ignored can be found in the supporting information.
    \item $^{**}$ We omit the results of SwimOCSR on the large datasets that take it more than one day to complete.
    \end{tablenotes}
    \end{threeparttable}
    }
}
\end{table*}

Following the setup of \citeauthor{img2mol}, we further evaluate the model on perturbed datasets with slight image rotation and shearing. \rev{\ours continues to exhibit robust performance on these datasets, outperforming other methods by large margins.} MolVec and OSRA's performance drops significantly as compared to the performance on original images, showing that rule-based systems are sensitive to input perturbations. \rev{We note that our baseline model performs slightly better than \ours on UOB$_p$ and Staker$_p$. It is because these two datasets do not contain chiral molecules and the strength of \ours is concealed. The evaluation in \Cref{tab:main} requires the chirality in the predicted molecules to be correct. It can be unfair for Img2Mol, which does not predict chirality. In the supporting information, we present additional evaluation results when chirality is ignored.  
}


\begin{figure}[t]
    \centering
    \includegraphics[width=\linewidth]{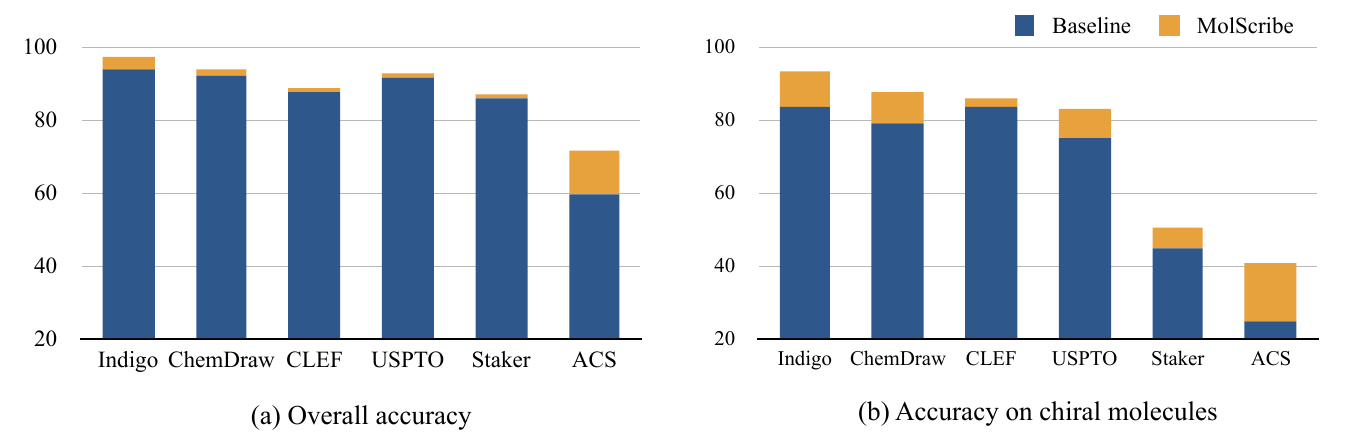}
    \caption{\ours explicitly determines chirality from the predicted graph and coordinates, thus improving recognition accuracy on chiral molecules.}
    \label{fig:chiral_acc}
\end{figure}

\paragraph{Chirality}
\Cref{fig:chiral_acc} compares the performance of our baseline image-to-SMILES model and \ours on chiral molecules.
\ours predicts chirality more accurately than the baseline. \rev{As discussed in the Stereochemistry section, end-to-end neural models make mistakes more frequently on chiral molecules, as it requires geometric reasoning over the graph structure to determine the chirality. \ours, on the other hand, predicts all the atom coordinates and bond types and explicitly determines the chiral types. This design leads to significant improvement on chiral molecules.}

\begin{table}[t]
    \centering
    \caption{Ablation study of \ours. We train model variants with 200K synthetic images generated by Indigo, and evaluate their in-domain and out-of-domain performance.}
    \label{tab:ablation}
    \begin{tabular}{lcccccccc}
    \toprule
        & \makecell{Indigo\\\footnotesize(in-domain)}  & \makecell{ChemDraw\\\footnotesize(out-of-domain)} \\\midrule
    Baseline & 88.7 & 82.5  \\
    \quad - without data augmentation & 86.0 & 57.2 \\\midrule
    \ours & 95.9 & 89.7 \\
    \quad - without data augmentation & 91.4 & 72.1 \\
    \quad - continuous coordinates & 89.3 & 79.5 \\
    \quad - remove non-atom tokens & 95.3 & 89.4 \\
    \bottomrule
    \end{tabular}
\end{table}

\rev{\paragraph{Ablation Study}
\Cref{tab:ablation} shows the ablation study of our model design. In this experiment, we use a subset of our synthetic training data (200K images) and train three model variants to validate the effects of several components of \ours. } First, the model trained without data augmentation achieves much worse performance on the out-of-domain ChemDraw dataset, which consists of images generated by a different software than the one used to generate training data (Indigo). This demonstrates that our data augmentation strategy is the key for the model to generalize to different styles. 
Second, we train a model that predicts continuous atom coordinates with an additional layer on top of the decoder as a regression task~\cite{image2graph}. \ours performs better by converting the coordinates to discrete tokens and generating the atoms and coordinates in a sequence. Third, we remove the non-atom tokens (parentheses, digits, equal signs, etc.) from the decoder output, and the performance drops slightly. The non-atom tokens in SMILES indicate branching and connections, thus helping the decoder to better model the graph structure.

\begin{figure}[t]
    \centering
    \includegraphics[width=0.99\linewidth]{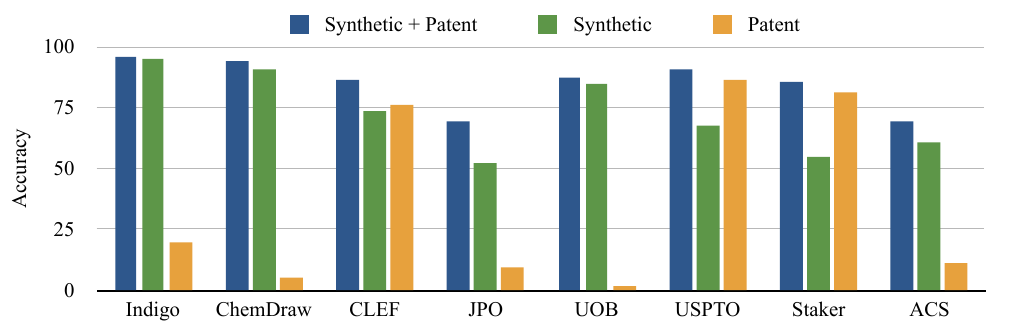}
    \caption{\ours is trained with both synthetic (PubChem) and patent (USPTO) data, and achieves better performance than training on each of them alone.}
    \label{fig:synreal}
\end{figure}

\Cref{fig:synreal} shows the contributions of the two datasets used to train \ours.
The model trained with patent data alone performs well on CLEF, USPTO, and Staker, which are also collected from patent images, but performs unsatisfactorily in other domains. Our synthetic data and data augmentation strategies contribute to \ours's robust performance across different domains.

\begin{figure}[t]
    \centering
    \begin{subfigure}[b]{0.4\textwidth}
        \centering
        \includegraphics[height=4.2in,clip,trim={20 0 170 0}]{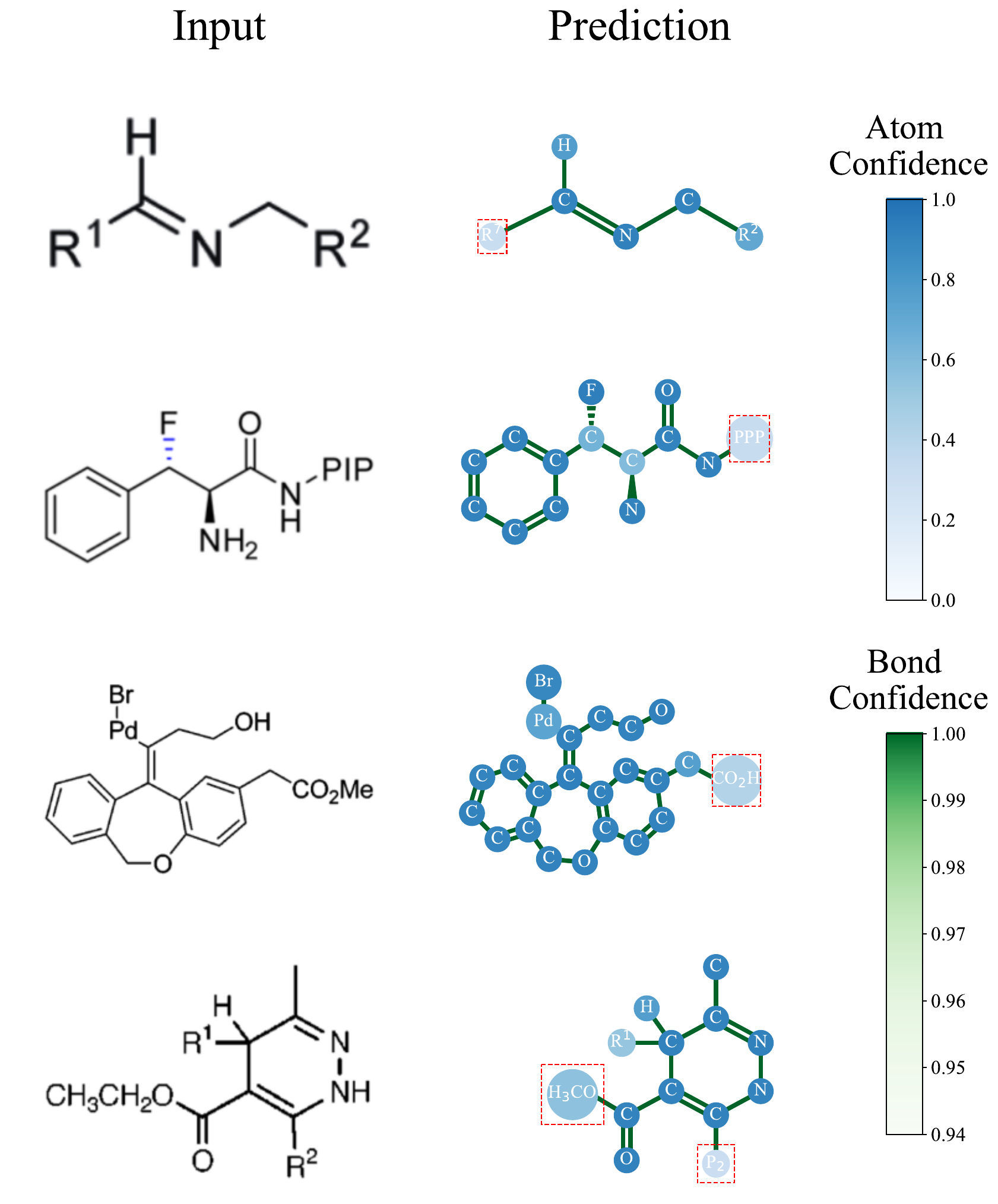}
        \caption{Incorrect atoms}
        \label{fig:aberr-atom}
    \end{subfigure}
    \hfill
    \begin{subfigure}[b]{0.52\textwidth}
        \centering
        \includegraphics[height=4.2in,clip,trim={20 0 0 0}]{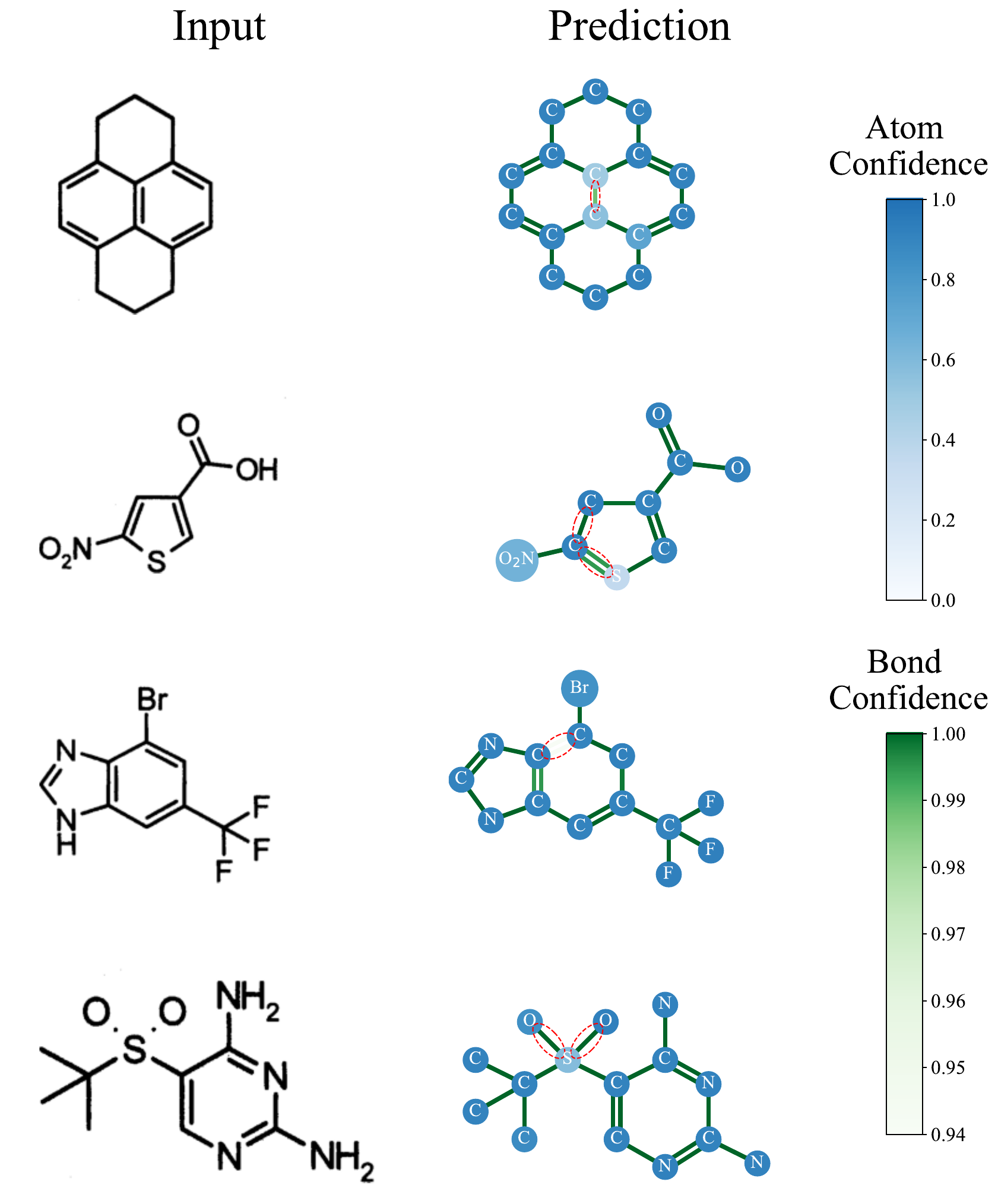}
        \caption{Incorrect bonds}
        \label{fig:aberr-bond}
    \end{subfigure}
    \caption{Examples of incorrectly predicted atoms/bonds and their confidence. Atom/bond confidence is represented by shade (darker means more confident), and mistakes are indicated with dashed boxes. }
    \label{fig:confidence}
\end{figure}

\begin{figure}[t]
    \centering
    \begin{subfigure}[b]{\linewidth}
        \includegraphics[width=0.95\linewidth]{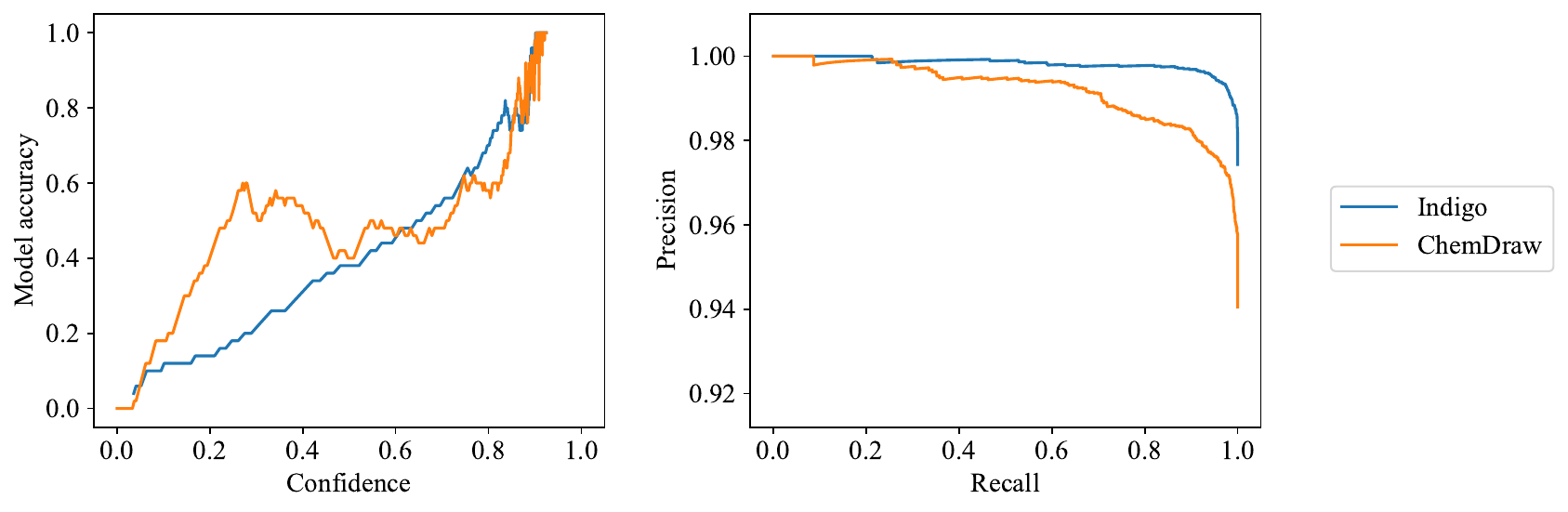}
        \caption{Synthetic Benchmarks}
    \end{subfigure}
    \par\bigskip
    \begin{subfigure}[b]{\linewidth}
        \includegraphics[width=0.95\linewidth]{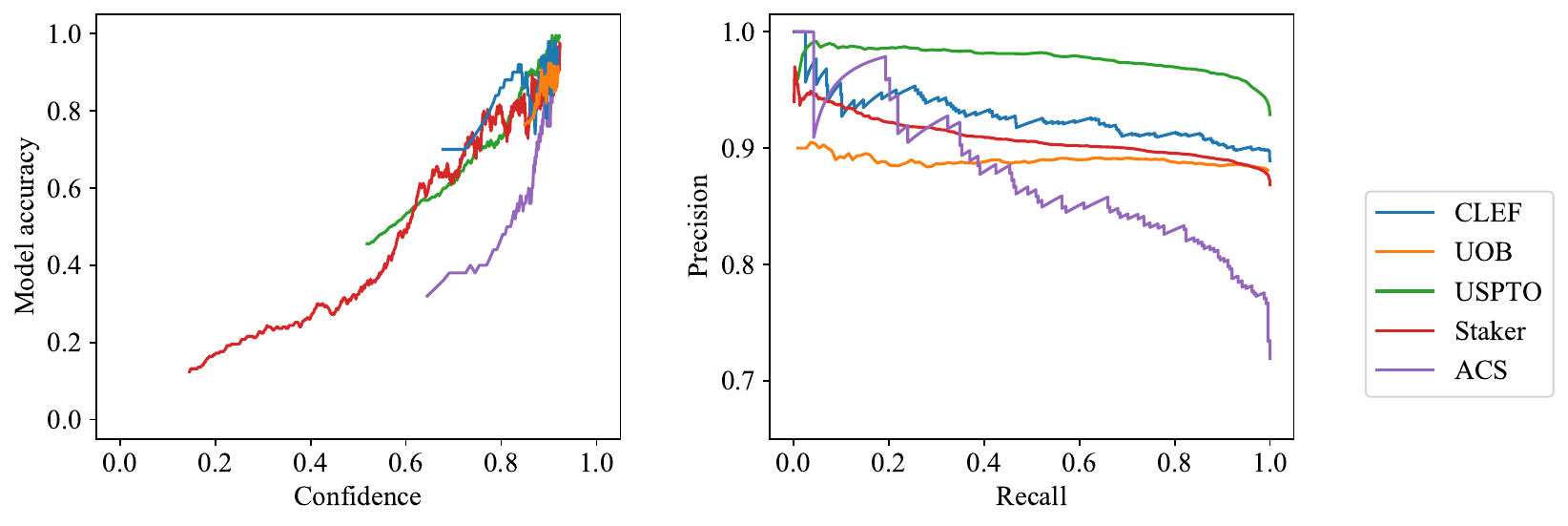}
        \caption{Realistic Benchmarks}
    \end{subfigure}
    \caption{\ours's confidence estimation on synthetic and realistic benchmarks. We show the correlation between model accuracy and confidence, and the precision-recall curve by setting thresholds on confidence.}
    \label{fig:prc}
\end{figure}

\paragraph{Model Confidence}

\ours can provide the confidence of its atom and bond predictions. \Cref{fig:confidence} shows a qualitative analysis of the model-assigned probabilities of atoms and bonds and their correctness.  Usually, lower probabilities are assigned to incorrect atoms/bonds, indicating the model is uncertain about certain parts of the image.

\ours also estimates how likely the entire molecular graph is correct by aggregating atom and bond confidence. We compute the molecule confidence with the average log probabilities of its atoms and bonds.
Figure \ref{fig:prc} shows the molecule-level confidence-accuracy and precision-recall curves. The precision-recall curve slopes downward and model accuracy increases with the confidence. By setting thresholds on confidence scores, we can obtain more confident and accurate predictions.   

\paragraph{Human Evaluation}
\ours also has the advantage of being more interpretable due to the atom-level alignment between the predicted graph and the molecular image. Given the molecular graph predicted by our model, chemists can easily determine whether it is the same molecule as depicted in the image and make corrections if there are mistakes. To validate this function, we asked three students with a chemistry background to conduct an experiment of converting molecular images to SMILES strings. We compare three setups:
\begin{enumerate}[(1)]
    \item Image-only: the students are given only the image and reconstruct the molecule;
    \item Predicted SMILES: the students are given the image and the predicted SMILES;
    \item Predicted graph: the students are given the image and the predicted molecular graph.
\end{enumerate}
As it is difficult to manually write the SMILES, the students use ChemDraw to edit the molecule's structure. In (2) and (3), the students import the prediction into ChemDraw, judge whether the prediction is correct, and edit it if there are any mistakes. Setup (2) (predicted SMILES) does not preserve the molecular layout, making it harder to compare. Each setup contains 20 images of similarly-sized molecules (34--36 atoms on average).  We record the time taken by the students to make the judgment and edit.

\begin{figure}[t]
    \centering
    \includegraphics[width=\linewidth]{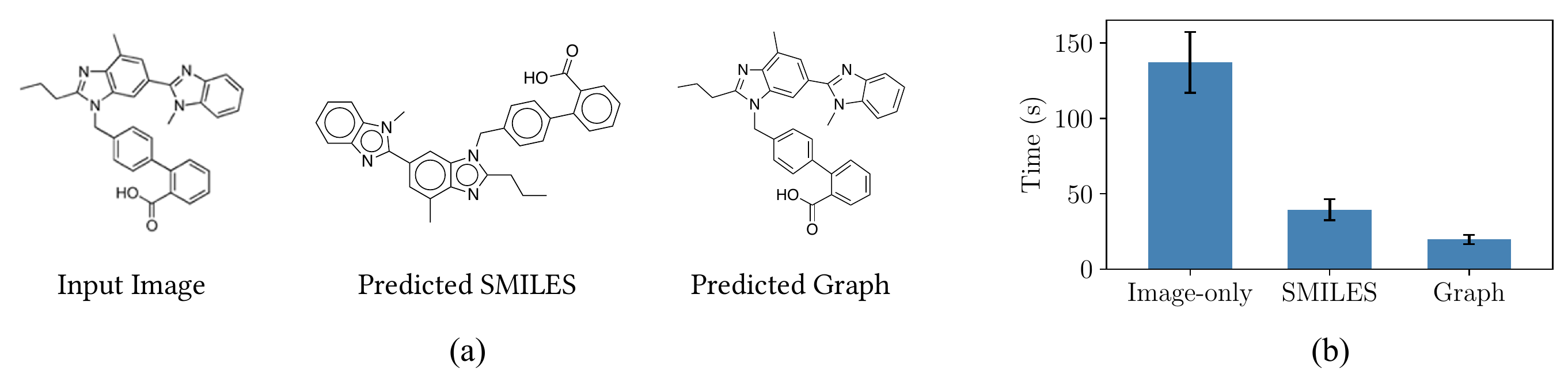}
    \caption{Human evaluation. (a) shows an example image with its predicted SMILES and molecular graph (imported into ChemDraw). (b) shows the average time taken by a student to convert a molecular image into SMILES, with or without the model prediction.}
    \label{fig:humaneval}
\end{figure}


\Cref{fig:humaneval} shows the experimental results. In the image-only setup, it took these students 137 seconds on average  to draw the depicted molecule in ChemDraw. (Note that these students are not power users of ChemDraw who can take advantage of the large number of keyboard shortcuts.) When the predicted SMILES is provided, the average time reduces to 39 seconds. When the predicted graph is provided, it further reduces to 20 seconds. The results clearly demonstrate the benefits of our graph generation method in visually inspecting the correctness of the model-generated structure.

%% file: conclusion.tex
\section{Conclusion}
In this paper, we propose \ours, an image-to-graph generation model for molecular structure recognition. \ours is built on an encoder-decoder architecture and jointly predicts atoms and bonds, along with their geometric layout. We design data augmentation strategies such that the model is robust to the domain shift and diverse patterns in real-world molecular images. Our model's flexibility allows us to incorporate symbolic chemistry constraints such as chirality verification and abbreviated structure expansion. Evaluations on both synthetic and realistic benchmarks show strong performance, surpassing existing rule-based and learned systems. Finally, our model produces more interpretable output, allowing chemists to easily examine the prediction and correct the mistakes. We release our code, data, and model for reproducibility, and provide interfaces so that chemists can easily use \ours in their research.

The scope of this work is limited to the recognition of single-molecule images. \ours achieves strong and robust performance on this task, which is an essential component in building a chemistry information extraction system\cite{chemschematicresolver,reactiondataextractor,guo2021automated}. \rev{An important future direction is to extend \ours to handle hand-drawn molecules.} \ours can recognize basic Markush structures with R-groups, but more complicated cases, such as the R-group that could be attached to any atom of a ring or a variable number of repetitions of a substructure, are not covered. Such cases cannot be represented as SMILES strings or MOLfiles, and future work may need to design a specialized format to handle Markush structures.  

\section{Data and Software Availability}
Our code, data, and model checkpoints are publicly available at \url{https://github.com/thomas0809/MolScribe}. We have also developed a web interface for \ours: \url{https://huggingface.co/spaces/yujieq/MolScribe}.  More details about the data sources can be found in the supporting information.